\pdfoutput=1

\documentclass[11pt]{article}

\usepackage[preprint]{acl}

\usepackage{times}
\usepackage{latexsym}

\usepackage[T1]{fontenc}

\usepackage[utf8]{inputenc}

\usepackage{microtype}

\usepackage{inconsolata}

\usepackage{graphicx}

\usepackage{booktabs}
\usepackage{algorithm}
\usepackage{makecell}
\usepackage{subfigure}
\usepackage{algorithmic}
\usepackage{amsmath}
\usepackage{multirow}
\usepackage{colortbl}
\usepackage{xcolor}
\usepackage{arydshln}
\usepackage{pdflscape}
\usepackage{rotating}

\definecolor{midnightgreen}{rgb}{0.0, 0.29, 0.33}

\newcommand{\ours}{\textsc{Craw4LLM}}
\title{\ours{}: Efficient Web Crawling for LLM Pretraining}

\author{Shi Yu$^{12}$\thanks{Work done while visiting Carnegie Mellon University.} \quad Zhiyuan Liu$^1$ \quad Chenyan Xiong$^2$
\\ $^{1}$Department of Computer Science and Technology, Tsinghua University\\
$^2$Language Technologies Institute, Carnegie Mellon University\\
\texttt{yus21@mails.tsinghua.edu.cn}; \texttt{liuzy@tsinghua.edu.cn};
\texttt{cx@cs.cmu.edu}
}

\begin{document}
\maketitle

\begin{abstract}
Web crawl is a main source of large language models' (LLMs) pretraining data, but the majority of crawled web pages are discarded in pretraining due to low data quality.
This paper presents \ours{}, an efficient web crawling method that explores the web graph based on the preference of LLM pretraining. 
Specifically, it leverages the influence of a webpage in LLM pretraining as the priority score of the web crawler's scheduler, replacing the standard graph-connectivity-based priority.
Our experiments on a web graph containing 900 million webpages from a commercial search engine's index demonstrate the efficiency of \ours{} in obtaining high-quality pretraining data.
With just 21\% URLs crawled,
LLMs pretrained on \ours{} data reach the same downstream performances of previous crawls, significantly reducing the crawling waste and alleviating the burdens on websites.
Our code is publicly available at \url{https://github.com/cxcscmu/Craw4LLM}.
\end{abstract}

\section{Introduction}

Massive in size and diverse in topics, web data usually serve as the primary source of pretraining data for large language models (LLMs), providing an extensive and heterogeneous corpus that captures a wide spectrum of human knowledge and real-world information~\citep{cc-analysis, dubey2024llama, fineweb}.
Pretraining datasets are typically built from large-scale web crawls such as Common Crawl~\citep{CommonCrawl2007}, which may contain TBs of data spanning billions of webpages~\citep{fineweb,redpajamav2}.

\begin{figure}[t]
    \centering
    \includegraphics[width=0.99\columnwidth]{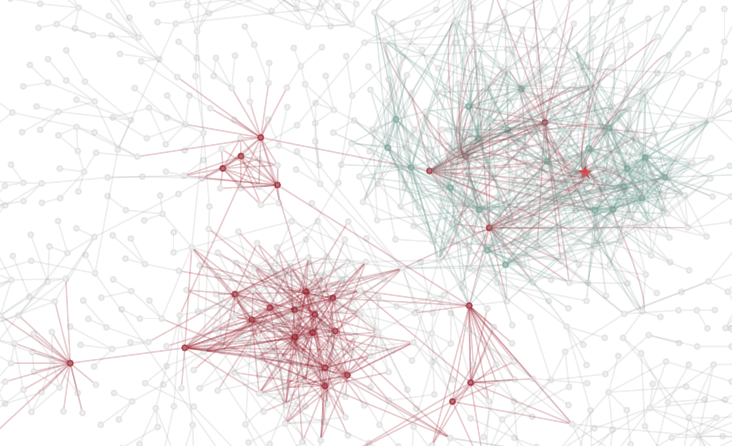}
    \caption{Graph traverse process of a traditional graph-connectivity-based crawler (green) and \ours{} (red) starting from a same seed URL (star). 
    }
    \label{fig:graph_traverse}
\end{figure}

Despite their vast scale, most of the data collected from web crawls are not used in the pretraining of LLMs.
Existing work often discards over 90\% of the raw data collected from the web~\citep{dclm,fineweb,txt360}, highlighting the \textit{inefficiency} of current web crawlers in collecting LLM pretraining data.
Common web crawlers like Common Crawl prioritize pages based on graph connectivity metrics like PageRank~\citep{pagerank,pr-crawl} or harmonic centrality~\citep{harmonic,cc-analysis}, which favor documents with a high number of inlinks (indegree)~\citep{pr-and-indegree} rather than those most relevant for pretraining.
This misalignment not only leads to waste in computational resources during excessive data processing for LLM developers, but also incentivizes over-crawling, which burdens website operators with redundant traffic and increases ethical and legal risks related to fair use of data and copyright~\citep{longpre2024consent,nyt_vs_openai}.

To bridge this gap, we propose Web \textbf{Craw}ling \textbf{for} \textbf{LL}\textbf{M} Pretraining (\ours{}).
Instead of relying on traditional graph-connectivity-based signals, \ours{} improves crawling efficiency by prioritizing webpages based on their influence on LLM pretraining.
Specifically, during each crawling iteration, all newly discovered documents are scored with a pretraining influence scorer derived from data-filtering pipelines for pretraining~\citep{dclm,fineweb}, and documents with the highest scores are used to discover new documents.
By prioritizing webpages with high influence scores, as illustrated in Figure~\ref{fig:graph_traverse}, \ours{} explores the web graph in a fundamentally different manner from traditional graph-connectivity-based crawlers, uncovering a distinct subset of the web more useful for pretraining.

We conduct large-scale crawling simulations on ClueWeb22-A~\citep{clueweb22}, a snapshot of the web containing 900 million English webpages obtained from the central index of a commercial search engine.
Results show that, by crawling only 1× of the pretraining dataset size, \ours{} can outperform traditional crawlers which collect 1×, 2×, and 4× data followed by data selection.
Compared to the baseline crawler that achieves the same performance, \ours{} crawls only 21\% of the webpages. 
Further analysis reveals that during crawling, \ours{} quickly discovers documents that align with the oracle selection, which selects from the full web graph. 
As a result, it achieves over 95\% of the oracle performance while crawling only 2.2\% of the data.

\section{Methodology}


\begin{figure}[t]
    \centering
    \subfigure[Pretraining (-0.11).\label{fig:joint_distribution:fasttext_indegree}]{
    \includegraphics[width=0.48\columnwidth]{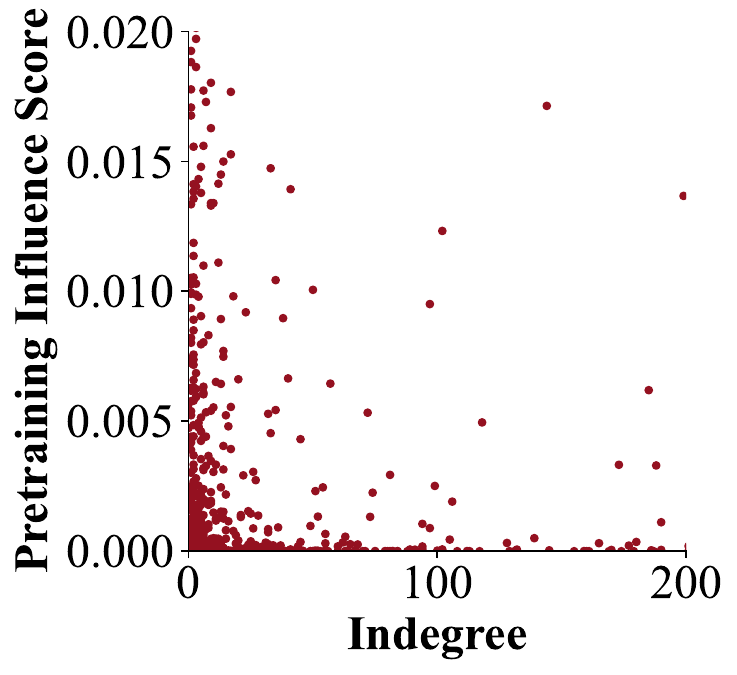}}
    \subfigure[PageRank (0.88).\label{fig:joint_distribution:indegree_pr}]{
    \includegraphics[width=0.48\columnwidth]{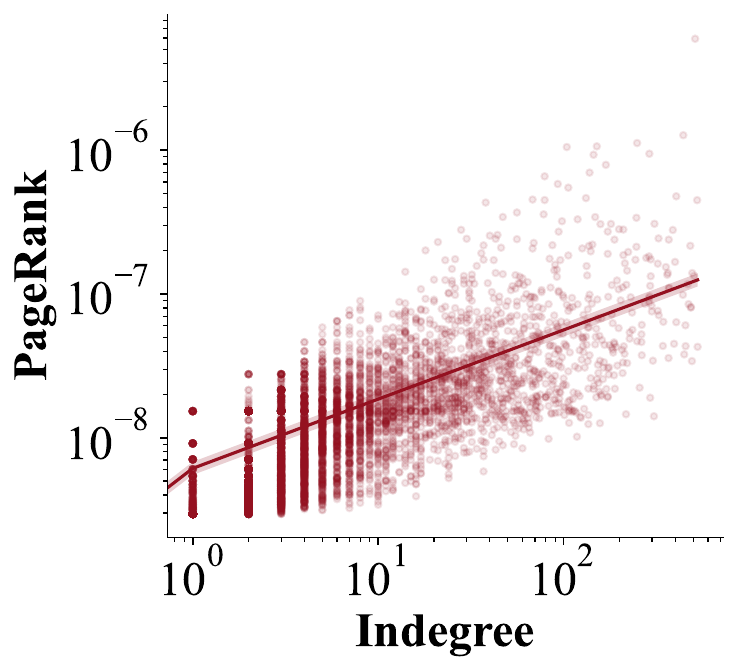}}
    \caption{Correlations between pretraining influence scores from DCLM fastText~\citep{dclm} and PageRank to indegrees, on randomly sampled  ClueWeb22-B documents~\citep{clueweb22}. Spearman correlation coefficients are reported in parentheses.}
    \label{fig:joint_distribution}
\end{figure} 

In this section, we introduce Web Data \textbf{Craw}ling \textbf{for} \textbf{LL}\textbf{M} Pretraining (\ours{}), an efficient crawling method that integrates LLM pretraining preference into the crawler.
The algorithm of \ours{} is presented in Algorithm~\ref{algo:crawl}.

\begin{algorithm}[!tb]
\small
\caption{\ours{} Algorithm}
\label{algo:crawl}
\begin{algorithmic}[1]
    \REQUIRE Seed URLs $\mathcal{U}_{\mathrm{seed}}$, number of pages to be crawled $N$, number of pages to be crawled in each iteration $n$, pretraining influence scorer $\mathcal{M}(\cdot)$
    \ENSURE Crawled page set $\mathcal{P}$

    \STATE Initialize \textbf{URL and score \textit{priority} queue} $\mathcal{Q} \gets \emptyset$
    \STATE Initialize \textbf{crawled page set} $\mathcal{P} \gets \emptyset$
    \STATE Initialize \textbf{visited URL set} $\mathcal{V} \gets \mathcal{U}_{\mathrm{seed}}$
    \STATE $\mathcal{U}_c \gets \mathcal{U}_{\mathrm{seed}}$

    \WHILE{$\lvert \mathcal{P} \rvert \leq N$}
        \STATE $\mathcal{P}_c \gets \textsc{FetchPages}(\mathcal{U}_c)$ 
        \STATE Merge $\mathcal{P}_c$ into $\mathcal{P}$
        \STATE $\mathcal{U}_{\mathrm{out}} \gets \textsc{ExtractURLs}(\mathcal{P}_c)$ 
        \FORALL{$v \in \mathcal{U}_{\mathrm{out}}$}
            \IF{$v \notin \mathcal{V}$}
                \STATE $\textsc{Enqueue}(\mathcal{Q},\; v, \;\textsc{Score\_URL}(v;\mathcal{M}))$ \label{line:score}
                \STATE $\textsc{Add}(\mathcal{V},\; v)$
            \ENDIF
        \ENDFOR
        \STATE $\mathcal{U}_c \gets \textsc{Dequeue}(\mathcal{Q},\; n)$
    \ENDWHILE

    \RETURN $\mathcal{P}$
\end{algorithmic}
\end{algorithm}

Similar to traditional crawlers~\citep{pr-crawl}, \ours{} starts with a set of seed URLs.
For each unvisited outlink of them, \ours{} assigns a score using a pretraining-oriented URL scoring function $\textsc{Score\_URL}(\cdot;\mathcal{M})$, where $\mathcal{M}$ is a pretraining influence scorer which rates a document's influence for pretraining.
$\mathcal{M}$ can be derived from data classification models for pretraining data, which have been used to decide whether a document should be retained in or filtered out from the raw dataset~\citep{dclm,fineweb}.
Formally, given a pretraining influence scorer $\mathcal{M}$, the score $s$ of a URL $u$ is calculated as
\begin{equation}
\label{eq:score}
\small
    s \gets \textsc{Score\_URL}(u;\mathcal{M}) = \mathcal{M}(\textsc{FetchPage}(u)),
\end{equation}
where $\textsc{FetchPage}(u)$ gets the page content of $u$ and $\mathcal{M}(\cdot)$ returns the score.
Once all outlinks have been scored, following the standard procedures of existing crawlers, they are inserted into a priority queue, which automatically orders them based on their scores.
The top $n$ highest-scoring URLs are then dequeued for pretraining and serve as the sources for the next round of crawling.
This process repeats until $N$ documents have been collected, forming the final pretraining dataset $\mathcal{P}$.

\begin{table*}[t]
  \centering
  \resizebox{1.0\textwidth}{!}{
    \begin{tabular}{rlccccccc}
    \toprule 
    & \makecell{\textbf{\ }\\\textbf{Selection}} & \makecell{\textbf{Commonsense}\\\textbf{Reasoning}} & \makecell{\textbf{Language}\\\textbf{Understanding}} & \makecell{\textbf{Reading}\\\textbf{Comprehension}} & \makecell{\textbf{Symbolic}\\\textbf{Problem Solving}} & \makecell{\textbf{World}\\\textbf{Knowledge}} & \makecell{\textbf{Core}} & \makecell{\\\textbf{\% of}}  \\

    \textbf{Method} & \textbf{Pool Size} & \textit{(4 tasks)} & \textit{(6 tasks)} & \textit{(3 tasks)} & \textit{(5 tasks)} & \textit{(5 tasks)} & \textit{(23 tasks)} & \textbf{Oracle}\\
    
    \midrule
    
    \multicolumn{9}{l}{Using the \textbf{DCLM fastText}~\citep{dclm} classifier as the pretraining influence scorer}\\
    
    Oracle & 45× & 0.2438 & 0.2209 & 0.1483 & 0.2039 & 0.2403 & 0.2239 & 100\% \\

    \hdashline

    
    \multirow{2}{*}{Random} & 1× & \underline{0.1906} & 0.1890 & 0.0244 & 0.1834 & 0.1930 & 0.1748 & 78.1\% \\
    
    & 2× & 0.1896 & 0.1967 & \textbf{0.1260} & \textbf{0.2000} & \underline{0.2024} & \underline{0.1964} & \underline{87.7\%}\\
    
    \multirow{2}{*}{Indegree} & 1× & 0.1730 & 0.1680 & 0.0326 & 0.1616 & 0.1668 & 0.1556 & 69.5\%\\
    
    & 2× & 0.1845 & 0.1856 & 0.0970 & 0.1958 & 0.1953 & 0.1865 & 83.3\%\\
    
    \hdashline
    
    
    \rowcolor{blue! 12} 
    \ours{} & 1× & \textbf{0.2116} & \textbf{0.2311} & 0.0826 & \underline{0.1979} & \textbf{0.2486} & \textbf{0.2133} & \textbf{95.3\%}\\
    \midrule
    
    \multicolumn{9}{l}{Using the \textbf{FineWeb-Edu}~\citep{fineweb} classifier as the pretraining influence scorer}\\
    
    Oracle & 45× & 0.1899 & 0.1973 & 0.1081 & 0.2117 & 0.2786 & 0.2133 & 100\% \\
    
    \hdashline
    
    
    \multirow{2}{*}{Random} & 1× & \underline{0.1906} & \textbf{0.1890} & 0.0244 & \underline{0.1834} & 0.1930 & 0.1748 & 82.0\% \\
    
    & 2× & 0.1797 & \underline{0.1888} & \textbf{0.0931} & 0.1586 & \underline{0.2100}  & \underline{0.1807} & \underline{84.7\%}\\
    
    \multirow{2}{*}{Indegree} & 1× & 0.1730 & 0.1680 & 0.0326 & 0.1616 & 0.1668 & 0.1556 & 72.9\%\\
    
    & 2× & 0.1720 & 0.1709 & \underline{0.0840} & 0.1783 & 0.1842 & 0.1724 & 80.8\%\\
    
    \hdashline
    
    
    \rowcolor{blue! 12} 
    \ours{} & 1× & \textbf{0.2122} & 0.1867 & 0.0368 & \textbf{0.2055} & \textbf{0.2837} & 0.2043 & \textbf{95.8\%} \\
    \bottomrule
  \end{tabular}
  }
  \caption{\label{tab:overall}
  Downstream LLM performance. 
  Either the DCLM fastText classifier~\citep{dclm} (top) or FineWeb-Edu classifier~\citep{fineweb} (down) is used as the pretraining influence scorer for \ours{}, and to select documents for oracle and crawl-then-select runs (Section~\ref{sec:experimental_method}).
  All models are pretrained on 1× data (20M documents, 32.9B tokens).
  The evaluation metric is centered accuracy (0 = random guess)~\citep{dclm}.
  Best/2nd best in the last two groups are bolded/underlined.
  See Appendix~\ref{sec:appendix:result} for detailed results.
  }
\end{table*}

In contrast, traditional crawlers typically rely on graph connectivity metrics, such as PageRank~\citep{pr-crawl} and harmonic centrality~\citep{cc-analysis}, which basically assign higher priority to pages with higher indegrees~\citep{pr-and-indegree}.
As shown in Figure~\ref{fig:joint_distribution:fasttext_indegree}, the indegrees of webpages exhibit a poor correlation with the scores assigned by the DCLM fastText classifier, a pretraining influence scorer for identifying high-quality pretraining data~\citep{dclm}.
This confirms that graph connectivity-based crawlers are inefficient in crawling pretraining data. 

By incorporating a pretraining influence scorer, \ours{} traverses the web graph in a way that prioritizes high-quality pretraining documents.  
This makes the crawling more efficient and enables the discovery of documents dramatically different with connectivity-based crawlers.

\section{Experimental Methodology}
\label{sec:experimental_method}

In this section, we introduce our experimental setup, with details on the crawler implementation and LLM training provided in Appendix~\ref{sec:appendix:crawl} and~\ref{sec:appendix:llm}.

\paragraph{\ours{}.}
To run experiments in our limited computational budget, we run a simulation of \ours{} on the ClueWeb22 dataset~\citep{clueweb22}, a snapshot of the web with graph information from a commercial crawler.
We use the English subset of ClueWeb22-A, which is a web graph containing 900M webpages with links.
We randomly sampled 10K URLs as our seed URLs.
We set the number of total crawled documents $N$ to 20M and crawled documents each iteration $n$ to 10K.
In our experiments, we consider using the DCLM fastText classifier~\citep{dclm} or the FineWeb-Edu classifier~\citep{fineweb} as the pretraining influence scorer $\mathcal{M}(\cdot)$, whose details can be found in Appendix~\ref{sec:appendix:scorers}.

\paragraph{Baselines.} 
We emulate traditional graph-connectivity-based crawlers by replacing the LLM-oriented URL scoring function (Eq.~\ref{eq:score}) with a function that returns the indegree for a given URL, since a node's indegree closely correlates with PageRank, a common graph connectivity metric, as shown in Figure~\ref{fig:joint_distribution:indegree_pr} and previous findings~\citep{pr-and-indegree}.
We also introduce a random crawling baseline, where the scorer assigns random scores.
We run both of them in a \textit{crawl-then-select} setting, first crawling 1× or 2× documents and then selecting the top 1× (20M) documents based on scores assigned by the DCLM fastText classifier or the FineWeb-Edu classifier.
This process mimics existing data-filtering pipelines, which begin with crawled documents and then apply filtering~\citep{dclm,fineweb}.

\paragraph{Oracle.}
We introduce an oracle selection run in which we directly apply the pretraining influence scorer to the ClueWeb22-A dataset and randomly sample 20M documents which are scored as the 10\% for pretraining, serving as the upper bound.

\paragraph{LLM Training and Evaluation.} 
For all runs, we use the final set of 20M crawled or selected documents to pretrain a 411M Transformer on 4× Chinchilla-optimal tokens~\citep{chinchilla}, totaling 32.9B tokens. 
The pretraining is conducted using the DCLM codebase~\citep{dclm}. 
To evaluate the pretrained LLMs, we follow the DCLM evaluation recipe, assessing performance on 23 (22 unique) \textit{core} tasks. 


\section{Evaluation Results}

In this section, we first present the overall performance of \ours{} (Section~\ref{sec:result:overall}), followed by further analysis (Section~\ref{sec:result:analysis}).

\subsection{Overall Performance}
\label{sec:result:overall}

In this experiment, we compare the performance of \ours{} with baseline crawlers by evaluating LLMs trained on their respective crawled data.
As shown in Table~\ref{tab:overall}, when all methods crawl the same amount of training data (1×), \ours{} significantly outperforms random crawling and indegree crawling using either DCLM fastText or FineWeb-Edu classifier as the pretraining influence scorer.
In the crawl-then-select setting, where traditional crawlers are allowed to collect twice as much data (2×) for later selection, they still underperform compared to \ours{}.
This suggests that incorporating pretraining-oriented signals early in the crawling process is more beneficial than relying on post-selection.
With only 1× of the data, \ours{} retains 95.3\% and 95.6\% of the performance achieved by the DCLM fastText and FineWeb-Edu oracle run, respectively, which directly selects from the entire ClueWeb22-A dataset, a substantially larger 45× data pool. 

\subsection{Analysis}
\label{sec:result:analysis}

In this subsection, we further analyze the efficiency of \ours{} compared to traditional crawlers and explore the reasons behind it.
We utilize the DCLM fastText classifier in the experiments presented in this subsection.

\begin{figure}[t]
    \centering
    \subfigure[Extended crawling.\label{fig:crawl_efficiency:pool}]{
    \includegraphics[width=0.48\columnwidth]{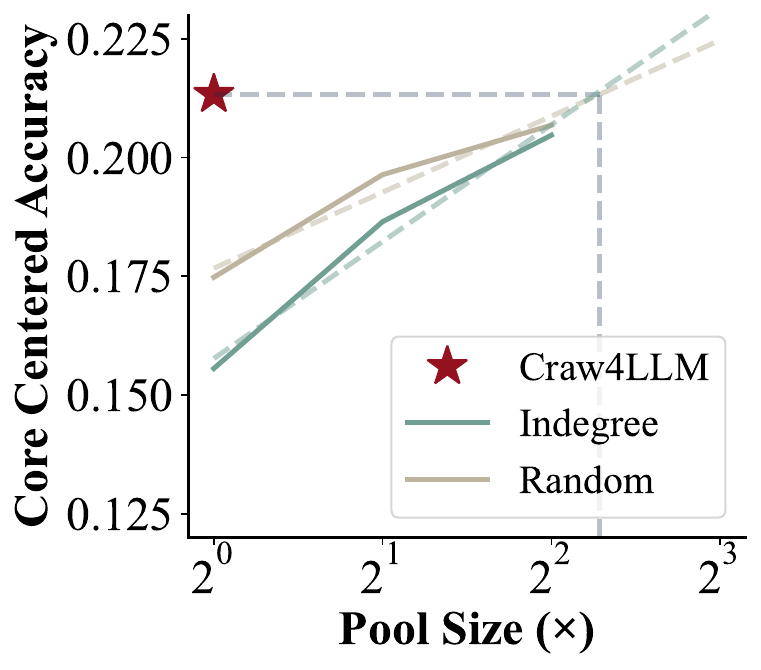}}
    \subfigure[Visited documents.\label{fig:crawl_efficiency:visited_document}]{
    \includegraphics[width=0.48\columnwidth]{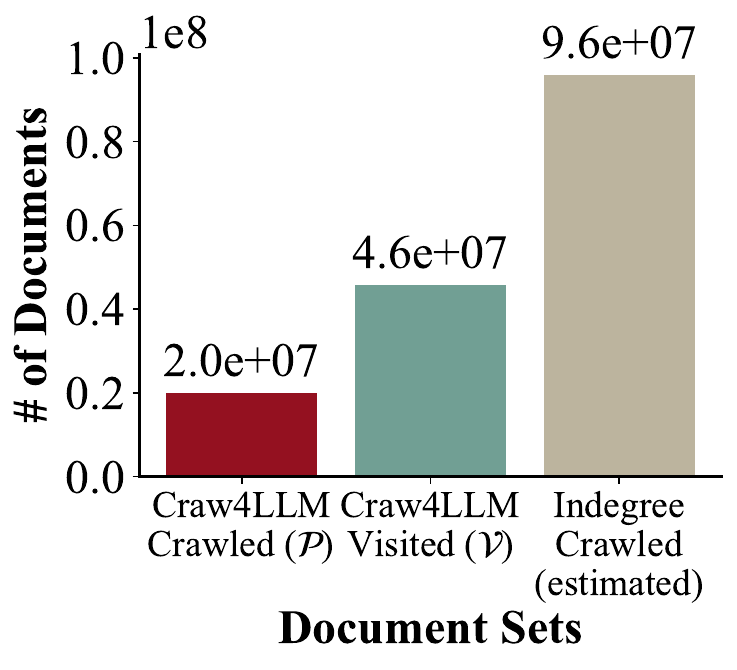}}
    \caption{Efficiency of crawlers. (a) shows the performance of LLMs trained on selected data crawled by \ours{} and extended baseline crawlers. (b) presents the number of crawled ($\mathcal{P}$) and visited ($\mathcal{V}$) documents for \ours{}, along with the estimated number of crawled documents required for indegree-based crawler to match \ours{}’s performance. 
    }
    \label{fig:crawl_efficiency}
\end{figure}

\paragraph{Crawling Efficiency.}
We evaluate the efficiency of \ours{} by comparing the number of documents it crawls or visits against baseline crawlers. 
As shown in Figure~\ref{fig:crawl_efficiency:pool}, even when the baselines crawl 4× the required pretraining data for selection, they still underperform compared to \ours{}. 
Extrapolation suggests that the indegree-based crawler would need to crawl 4.8× documents (96M) to match \ours{}’s performance. 
Figure~\ref{fig:crawl_efficiency:visited_document} further illustrates that \ours{} achieves the same performance while crawling only 21\% of the documents required by the indegree-based crawler, or 48\% when considering all visited documents.
These results highlight the efficiency of \ours{}, demonstrating its potential to reduce website burdens and mitigate over-crawling.

\paragraph{Document Coverage.}


\begin{figure}[t]
    \centering
    \includegraphics[width=0.99\columnwidth]{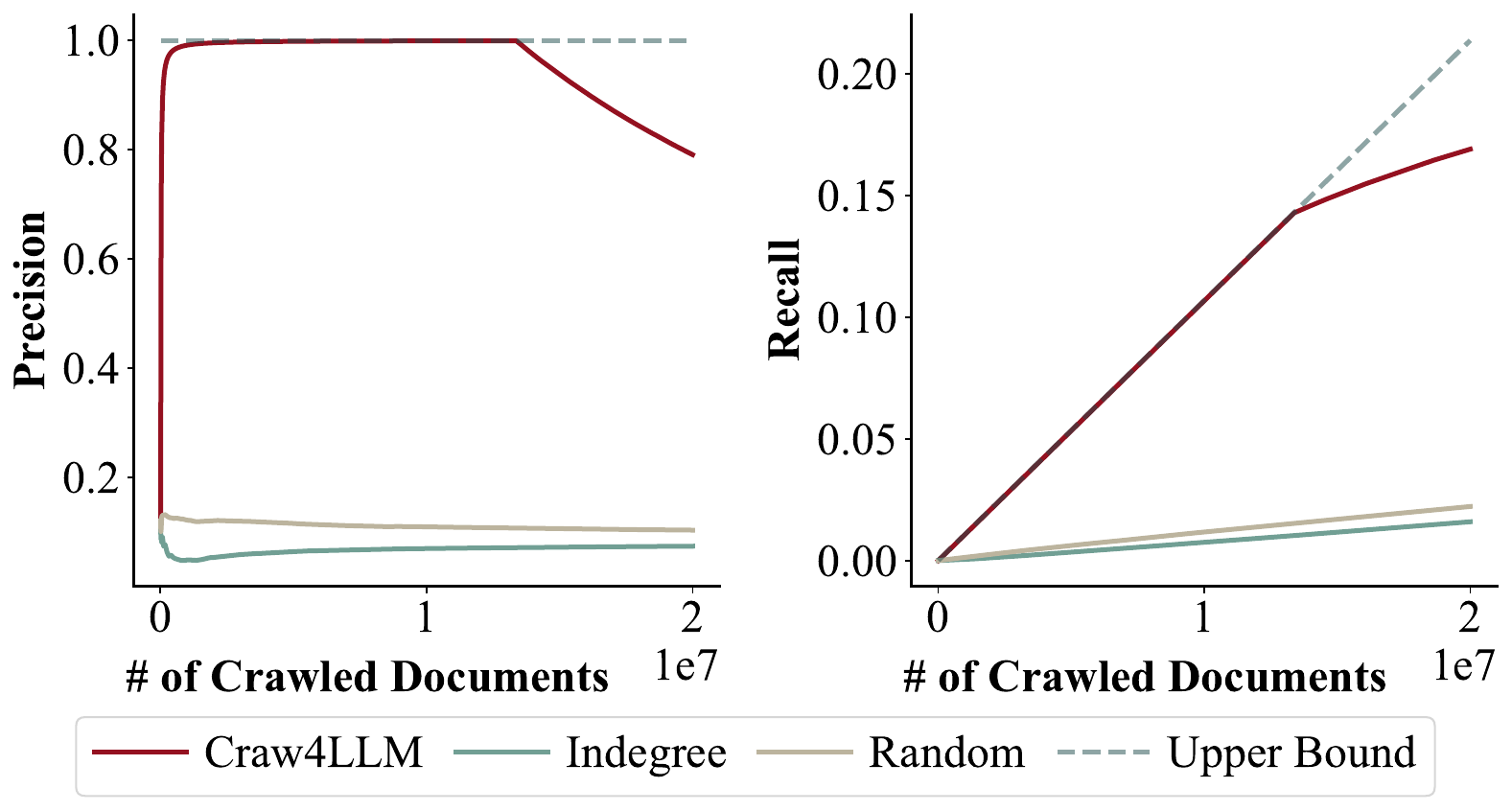}
    \caption{Precision (left) and recall (right) of the oracle documents among the documents crawled by \ours{}, indegree, and random crawler. The upper bound represents always crawling the oracle documents.}
    \label{fig:overlap}
\end{figure}

\begin{figure}[t]
    \centering
    \subfigure[1-hop (0.61).]{
    \includegraphics[width=0.48\columnwidth]{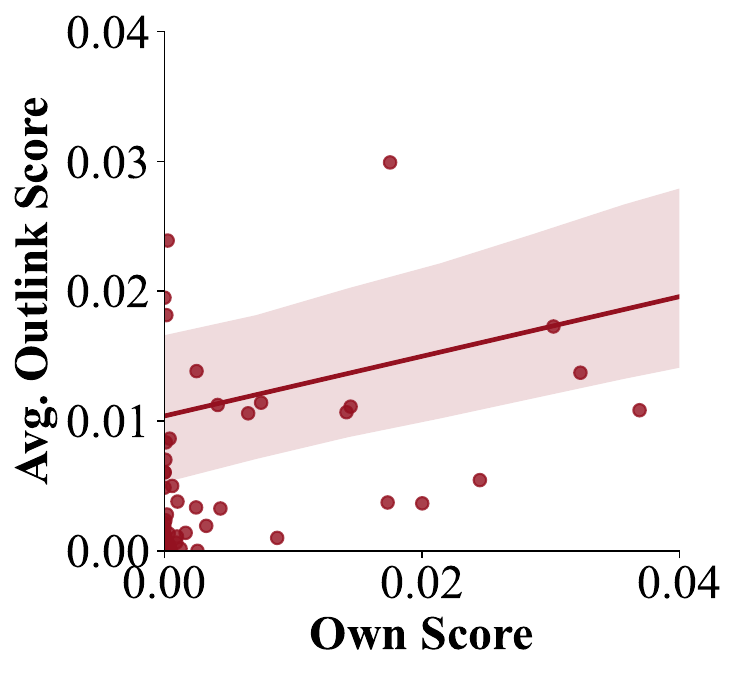}}
    \subfigure[2-hop (0.60).]{
    \includegraphics[width=0.48\columnwidth]{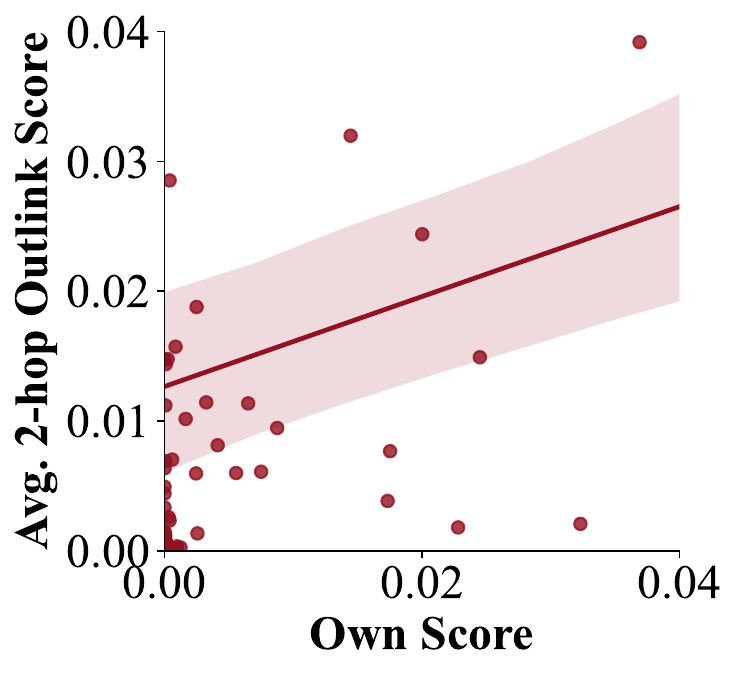}}
    \caption{Correlations between the pretraining influence scores of the documents themselves and the average scores of their 1- and 2-hop outlink documents.
    Spearman correlation coefficients are reported in parentheses.
    }
    \label{fig:correlation}
\end{figure}

In this experiment, we plot the precision and recall of the oracle-selected documents among those crawled by \ours{} and baseline crawlers throughout the crawling process.
As shown in Figure~\ref{fig:overlap}, the precision quickly reaches 1.0, while the recall increases linearly, aligning with the theoretical upper bound.
The saturated performance remains until 13 million documents have been crawled, after which the performance starts to decline, likely due to the lack of connectivity of the ClueWeb22 subgraph.
In contrast, baseline crawlers exhibit minimal overlap with oracle-selected data, verifying that most of their crawled content is misaligned with pretraining needs and should be filtered~\citep{dclm,fineweb}. 
These results emphasize the importance of targeted crawling strategies for pretraining.

\paragraph{Score Correlations Across Links.}

\ours{} tracks the outlinks of the highest-scored documents in the current iteration to enrich the queue for future crawls.
As shown in Figure~\ref{fig:correlation}, we plot the correlations between the pretraining influence scores of current documents and their 1- and 2-hop outlinks.
The results indicate a correlation in influence scores across link hops, suggesting that highly-rated documents are interconnected and can be discovered through previously crawled documents.

\section{Conclusion}

This paper presents \ours{}, a step toward more efficient and responsible web crawling for LLM pretraining. 
By prioritizing documents based on the pretraining needs, our method improves crawling efficiency and reduces unnecessary crawling, easing the burden on web hosts.
While fair use of web data remains a critical challenge, we hope that \ours{} can help mitigate these concerns and promote more compliant and sustainable practices in obtaining pretraining data for LLMs.

\section*{Limitations}

Web crawling raises important concerns regarding copyright and the fair use of web data~\citep{longpre2024consent}, necessitating a better solution from the entire LLM community, such as sharing benefits with website owners.
In this paper, we propose a more efficient crawling method that mitigates these challenges by reducing crawling, though it does not fully resolve them. 
Our experiments are conducted on a web graph dataset ClueWeb22~\citep{clueweb22}, thereby avoiding issues associated with actual web crawling. 
We hope that future advancements in web crawling will better align with ethical and legal standards.

While our crawling simulation is a sufficient research setup, further validation is required to assess the effectiveness of \ours{} in real-world crawling scenarios.  
Our \ours{} and baseline crawlers implement only the selection policy~\citep{pr-crawl} of a crawler, which determines which pages to crawl.  
Although we try to mimic real-world crawling procedures used in systems like Apache Nutch\footnote{\url{https://nutch.apache.org/}}, we do not implement other web crawling policies in industrial-level crawlers, such as the re-visit policy~\citep{revisit}, politeness policy~\citep{politeness}, and parallelization policy~\citep{parallelization}.  
We leave the integration of \ours{} into real-world crawling engines like Nutch and a comprehensive comparison between \ours{} and traditional crawling methods in real-world crawling scenarios for future work.

\bibliography{custom}

\appendix

\section{Details on Crawling}
\label{sec:appendix:crawl}

Our implementation of the indegree-based crawler employs a static URL scoring function, which directly returns the indegree of a given URL based on the full ClueWeb22 graph (Sec.~\ref{sec:experimental_method}).  
For real-world crawlers, as the true indegree value of a URL cannot be known in advance, a local graph must be maintained to track the \textit{known} inlinks of discovered URLs.  
This local graph is updated iteratively as the discovered portion of the web expands during the crawling process~\citep{pr-crawl}.  

Maintaining such a local graph during crawling introduces significant computational overhead.  
For simplicity, we instead implement the static simulation, where we directly return the global indegree of each URL.  
We believe that this simplified implementation does not underperform compared to real-world implementations, as our approach leverages global information from the entire graph, which should be better than the partial information from the local graph.

We run our simulated crawlers on a Linux server equipped with two Intel(R) Xeon(R) E5-2630 v3 CPUs (8 cores per socket, 16 cores in total, 1 thread per core), 125GiB of memory, and an SSD.  
The crawling process of \ours{} takes about 21 hours to finish.
In comparison, the random and indegree-based crawlers take around 10.5 and 12.5 hours, respectively.
Note that since these are simulated crawls on the snapshot, the reported times do not reflect real-world crawling performance.

\section{Details on LLM Training and Evaluation}
\label{sec:appendix:llm}

\begin{table}
    \centering
    \begin{tabular}{rl}
    \toprule
        \textbf{Hyper-parameter} & \textbf{Value}\\
    \midrule
         $n_{\text{layers}}$ & 24 \\
         $n_{\text{heads}}$& 8\\
         $d_{\text{model}}$& 1,024\\
         $d_{\text{head}}$& 128\\
         Warmup& 2,000\\
         Learning Rate& 3e-3\\
         Weight Decay& 0.033\\
         z-loss& 1e-4\\
         Global Batch Size & 512 \\
         Sequence Length & 2048 \\
    \bottomrule
    \end{tabular}
    \caption{Model and training hyper-parameters.
    $n_{\text{layers}}$, $n_{\text{layers}}$, $d_{\text{model}}$, and $d_{\text{head}}$ denote the number of layers, attention heads, width, and width per attention head, respectively.
    }
    \label{tab:hyperparameters}
\end{table}

We pretrain a 411M-parameter\footnote{Sometimes referred to as 400M in the DCLM paper~\citep{dclm}.} decoder-only Transformer model using the DCLM training recipe~\citep{dclm}\footnote{\url{https://github.com/mlfoundations/dclm}}.
The hyper-parameters are presented in Tabel~\ref{tab:hyperparameters}.
To enhance training stability, we extend the original 411M-1x setting to 411M-4x, meaning the model is trained on 4 times the Chinchilla-optimal number of tokens~\citep{chinchilla}, which amounts to 32.9B tokens. 
The training process takes 1 day and 12 hours on 8 NVIDIA L40S GPUs.
For further details, please refer to the DCLM paper~\citep{dclm}.
Due to computational constraints, each pretraining experiment is conducted only once.

We use the DCLM evaluation recipe~\citep{dclm} to evaluate model performance on 23 (22 unique) \textit{core} tasks. 

\section{Details on the Pretraining Influence Scorers}
\label{sec:appendix:scorers}

\paragraph{DCLM fastText Classifier}\citep{dclm} is a quality filter based on fastText~\citep{joulin2017bag}, trained to distinguish between high- and low-quality pretraining data.
A dataset size of 400K examples (200K positive, 200K negative) is used for training. 
Negative samples are random documents from an earlier version of the RefinedWeb reproduction from the DCLM team.
Positive samples come from OpenHermes 2.5~\citep{oh25} and high-scoring posts from the r/ExplainLikeImFive (ELI5) subreddit.
The final DCLM-baseline dataset is created by selecting only the top 10\% of documents ranked by the classifier.

\paragraph{FineWeb-Edu Classifier}\citep{fineweb} is a model designed to identify and filter educational content within FineWeb. 
Based on the \texttt{Snowflake-arctic-embed-m}\footnote{\url{https://huggingface.co/Snowflake/snowflake-arctic-embed-m}} text embedding model, it is trained on 450,000 annotations generated by Llama-3-70B-Instruct, which assigned an educational quality score from 0 to 5 to every document. 
FineWeb-Edu is built by filtering out samples from FineWeb with scores lower than 3 with the trained classifier.


\section{Detailed Results}
\label{sec:appendix:result}

\begin{table*}
    \centering
    \begin{tabular}{rllcccc}
        \toprule
        \textbf{\ } & \textbf{Pretraining} & \textbf{Selection} & \multicolumn{4}{c}{\textbf{Commonsense Reasoning}} \\
        \textbf{Method} & \textbf{Influence Scorer} & \textbf{Pool Size} & CommonsenseQA & COPA & OpenBookQA & PIQA \\
        \midrule
        Oracle & DCLM fastText & 45× & 0.2850 & 0.7000 & 0.3300 & 0.6812 \\
        Oracle & FineWeb-Edu & 45× & 0.2342 & 0.6200 & 0.3620 & 0.6638 \\
        Random & -- & 1× & 0.2072 & 0.6700 & 0.2980 & 0.6746 \\
        Random & DCLM fastText & 2× & 0.2588 & 0.6200 & 0.3160 & 0.6785 \\
        Random & FineWeb-Edu & 2× & 0.2301 & 0.6200 & 0.3140 & 0.6779 \\
        Random & DCLM fastText & 4× & 0.2326 & 0.6400 & 0.3380 & 0.6757 \\
        Indegree & -- & 1× & 0.3219 & 0.6000 & 0.2780 & 0.6513 \\
        Indegree & DCLM fastText & 2× & 0.1966 & 0.6600 & 0.3040 & 0.6752 \\
        Indegree & FineWeb-Edu & 2× & 0.1974 & 0.6400 & 0.3160 & 0.6616 \\
        Indegree & DCLM fastText & 4× & 0.2088 & 0.6400 & 0.3400 & 0.6817 \\
        \ours{} & DCLM fastText & 1× & 0.2277 & 0.6600 & 0.3300 & 0.6926 \\
        \ours{} & FineWeb-Edu & 1× & 0.3219 & 0.6200 & 0.3500 & 0.6616 \\
        \bottomrule
    \end{tabular}
    \caption{Results for commonsense reasoning tasks.}
    \label{tab:commonsense_results}
\end{table*}

\begin{table*}
    \centering
    \resizebox{1.0\textwidth}{!}{
    \begin{tabular}{rllcccccc}
        \toprule
        \textbf{\ } & \textbf{Pretraining} & \textbf{Selection} & \multicolumn{6}{c}{\textbf{Language Understanding}} \\
        \textbf{Method} & \textbf{Influence Scorer} &\textbf{Pool Size} & BIG-Bench Lang. Id. & HellaSwag (zero-shot) & HellaSwag  & LAMBADA & Winograd & Winogrande \\
        \midrule
        Oracle & DCLM fastText & 45× & 0.2515 & 0.3856 & 0.3905 & 0.4432 & 0.6557 & 0.5130 \\
        Oracle & FineWeb-Edu & 45× & 0.2580 & 0.3831 & 0.3818 & 0.3643 & 0.6227 & 0.5185 \\
        Random & -- & 1× & 0.2490 & 0.3709 & 0.3716 & 0.3990 & 0.6044 & 0.5146 \\
        Random & DCLM fastText & 2× & 0.2468 & 0.3882 & 0.3925 & 0.4073 & 0.6007 & 0.5130 \\
        Random & FineWeb-Edu & 2× & 0.2485 & 0.3815 & 0.3800 & 0.3804 & 0.6007 & 0.5146 \\
        Random & DCLM fastText & 4× & 0.2521 & 0.4011 & 0.4019 & 0.4390 & 0.6154 & 0.5130 \\
        Indegree & -- & 1× & 0.2566 & 0.3515 & 0.3519 & 0.3596 & 0.5971 & 0.5004 \\
        Indegree & DCLM fastText & 2× & 0.2547 & 0.3749 & 0.3771 & 0.3773 & 0.5861 & 0.5241 \\
        Indegree & FineWeb-Edu & 2× & 0.2528 & 0.3636 & 0.3658 & 0.3672 & 0.5678 & 0.5193 \\
        Indegree & DCLM fastText & 4× & 0.2562 & 0.3994 & 0.4008 & 0.4159 & 0.6190 & 0.5178 \\
        \ours{} & DCLM fastText & 1× & 0.2544 & 0.4035 & 0.4048 & 0.4196 & 0.6593 & 0.5288 \\
        \ours{} & FineWeb-Edu & 1× & 0.2521 & 0.3726 & 0.3717 & 0.3478 & 0.6264 & 0.5083 \\
        \bottomrule
    \end{tabular}
    }
    \caption{Results for language understanding tasks.}
    \label{tab:language_results}
\end{table*}

\begin{table*}
    \centering
    \begin{tabular}{rllccc}
        \toprule
        \textbf{\ } & \textbf{Pretraining} & \textbf{Selection} & \multicolumn{3}{c}{\textbf{Reading Comprehension}} \\
        \textbf{Method} & \textbf{Influence Scorer} & \textbf{Pool Size} & BoolQ & CoQA  & SQuAD \\
        \midrule
        Oracle & DCLM fastText & 45× & 0.5755 & 0.2479 & 0.3139 \\
        Oracle & FineWeb-Edu & 45× & 0.5367 & 0.2305 & 0.3130 \\
        Random & -- & 1× & 0.5080 & 0.1799 & 0.1882 \\
        Random & DCLM fastText & 2× & 0.5807 & 0.2053 & 0.2759 \\
        Random & FineWeb-Edu & 2× & 0.5627 & 0.1997 & 0.2304 \\
        Random & DCLM fastText & 4× & 0.5911 & 0.2361 & 0.2951 \\
        Indegree & -- & 1× & 0.5324 & 0.1666 & 0.1616 \\
        Indegree & DCLM fastText & 2× & 0.5697 & 0.1843 & 0.2390 \\
        Indegree & FineWeb-Edu & 2× & 0.5670 & 0.1904 & 0.2011 \\
        Indegree & DCLM fastText & 4× & 0.5765 & 0.2147 & 0.2736 \\
        \ours{} & DCLM fastText & 1× & 0.5440 & 0.2264 & 0.2215 \\
        \ours{} & FineWeb-Edu & 1× & 0.4654 & 0.2146 & 0.3026 \\
        \bottomrule
    \end{tabular}
    \caption{Results for reading comprehension tasks.}
    \label{tab:reading_results}
\end{table*}

\begin{table*}
    \centering
    \resizebox{1.0\textwidth}{!}{
    \begin{tabular}{rllccccc}
        \toprule
        \textbf{\ } & \textbf{Pretraining} & \textbf{Selection} & \multicolumn{5}{c}{\textbf{Symbolic Problem Solving}} \\
        \textbf{Method} & \textbf{Influence Scorer} & \textbf{Pool Size} & AGI Eval LSAT-AR & BIG-Bench CS Algorithms  & BIG-Bench Dyck Lang.  & BIG-Bench Operators  & BIG-Bench Repeat Copy Logic \\
        \midrule
        Oracle & DCLM fastText & 45× & 0.2739 & 0.4341 & 0.2160 & 0.2143 & 0.0625 \\
        Oracle & FineWeb-Edu & 45× & 0.2826 & 0.4606 & 0.2870 & 0.1762 & 0.0313 \\
        Random & -- & 1× & 0.2391 & 0.4568 & 0.1970 & 0.2143 & 0.0000 \\
        Random & DCLM fastText & 2× & 0.2696 & 0.4538 & 0.2520 & 0.1762 & 0.0313 \\
        Random & FineWeb-Edu & 2× & 0.2000 & 0.4091 & 0.2050 & 0.1476 & 0.0313 \\
        Random & DCLM fastText & 4× & 0.1957 & 0.4568 & 0.2600 & 0.1857 & 0.0625 \\
        Indegree & -- & 1× & 0.2304 & 0.4371 & 0.1900 & 0.1429 & 0.0000 \\
        Indegree & DCLM fastText & 2× & 0.2609 & 0.4235 & 0.2340 & 0.2143 & 0.0313 \\
        Indegree & FineWeb-Edu & 2× & 0.2304 & 0.4545 & 0.2060 & 0.1619 & 0.0313 \\
        Indegree & DCLM fastText & 4× & 0.2174 & 0.4538 & 0.2530 & 0.1667 & 0.0938 \\
        \ours{} & DCLM fastText & 1× & 0.2696 & 0.4371 & 0.1620 & 0.2095 & 0.0938 \\
        \ours{} & FineWeb-Edu & 1× & 0.2609 & 0.4750 & 0.1780 & 0.2048 & 0.0938 \\
        \bottomrule
    \end{tabular}
    }
    \caption{Results for symbolic problem solving tasks.}
    \label{tab:symbolic_results}
\end{table*}

\begin{table*}
    \centering
    \resizebox{1.0\textwidth}{!}{
    \begin{tabular}{rllccccc}
        \toprule
        \textbf{\ } & \textbf{Pretraining} & \textbf{Selection} & \multicolumn{5}{c}{\textbf{World Knowledge}} \\
        \textbf{Method} & \textbf{Influence Scorer} & \textbf{Pool Size} & ARC Easy  & ARC Challenge  & BIG-Bench QA Wikidata  & Jeopardy  & MMLU \\
        \midrule
        Oracle & DCLM fastText & 45× & 0.5951 & 0.3166 & 0.4945 & 0.1176 & 0.2805 \\
        Oracle & FineWeb-Edu & 45× & 0.6343 & 0.3549 & 0.5407 & 0.1726 & 0.2706 \\
        Random & -- & 1× & 0.5152 & 0.2799 & 0.5186 & 0.0461 & 0.2552 \\
        Random & DCLM fastText & 2× & 0.5425 & 0.2807 & 0.5081 & 0.0648 & 0.2561 \\
        Random & FineWeb-Edu & 2× & 0.5535 & 0.2816 & 0.5194 & 0.0863 & 0.2483 \\
        Random & DCLM fastText & 4× & 0.5577 & 0.2867 & 0.5126 & 0.0970 & 0.2543 \\
        Indegree & -- & 1× & 0.4857 & 0.2509 & 0.4888 & 0.0138 & 0.2618 \\
        Indegree & DCLM fastText & 2× & 0.5248 & 0.2790 & 0.5205 & 0.0555 & 0.2464 \\
        Indegree & FineWeb-Edu & 2× & 0.5143 & 0.2679 & 0.5044 & 0.0553 & 0.2389 \\
        Indegree & DCLM fastText & 4× & 0.5749 & 0.2935 & 0.5084 & 0.0959 & 0.2430 \\
        \ours{} & DCLM fastText & 1× & 0.6103 & 0.3208 & 0.5143 & 0.1323 & 0.2661 \\
        \ours{} & FineWeb-Edu & 1× & 0.6427 & 0.3592 & 0.5319 & 0.1859 & 0.2736 \\
        \bottomrule
    \end{tabular}
    }
    \caption{Results for world knowledge tasks.}
    \label{tab:world_knowledge_results}
\end{table*}

The raw (uncentered) accuracy of all evaluation tasks is presented in Table~\ref{tab:commonsense_results}, \ref{tab:language_results}, \ref{tab:reading_results},
\ref{tab:symbolic_results}, and 
\ref{tab:world_knowledge_results}.
Please refer to~\citet{dclm} for more details on the evaluation tasks.


\section{The ClueWeb22 Dataset}

ClueWeb22~\citep{clueweb22} is distributed under a ``TREC-style'' license for research purpose.  
The dataset can be obtained by signing a data license agreement with Carnegie Mellon University\footnote{\url{https://lemurproject.org/clueweb22/obtain.php}}.
We use ClueWeb22 only for research purpose.

\section{Use of AI Assistants}

We use GitHub Copilot\footnote{\url{https://github.com/features/copilot}} to assist with coding and ChatGPT\footnote{\url{https://chatgpt.com/}} (powered by GPT-4o) to enhance the writing of this paper.

\end{document}